\documentclass[numbers,final,12pt]{elsarticle}

\usepackage{amssymb}
\usepackage{amsmath}

\usepackage{amsmath,amsfonts}
\usepackage{algorithmic}
\usepackage{array}
\usepackage[caption=false,font=normalsize,labelfont=sf,textfont=sf]{subfig}
\usepackage{textcomp}
\usepackage{stfloats}
\usepackage{url}
\usepackage{verbatim}
\usepackage{graphicx}
\usepackage{svg}
\usepackage{cleveref}
\usepackage{booktabs, multirow}
\usepackage{gensymb}
\usepackage{balance}
\usepackage{threeparttable}
\usepackage{array}        
\usepackage{arydshln}     

\usepackage{xcolor}
\usepackage{graphics} 
\usepackage{amsmath} 
\usepackage{amssymb}  
\usepackage{physics}  

\journal{Robotics and Autonomous Systems}

\begin{document}

\begin{frontmatter}



\title{Uncertainty Aware-Predictive Control Barrier Functions: Safer Human Robot Interaction through Probabilistic Motion Forecasting} 


\author[1]{Lorenzo Busellato\fnref{fn1}}
\ead{lorenzo.busellato@univr.it}
\author[1]{Federico Cunico\fnref{fn1,c}}
\ead{federico.cunico@univr.it}
\author[1]{Diego Dall'Alba}
\ead{diego.dallalba@univr.it}
\author[1]{Marco Emporio}
\ead{marco.emporio@univr.it}
\author[1]{Andrea Giachetti}
\ead{andrea.giachetti@univr.it}
\author[1]{Riccardo Muradore}
\ead{riccardo.muradore@univr.it}
\author[1]{Marco Cristani}
\ead{marco.cristani@univr.it}

\affiliation[1]{organization={University of Verona, Department of Engineering for Innovation Medicine},
    addressline={Strada Le Grazie, 15}, 
    city={Verona},
    postcode={37134}, 
    state={Veneto},
    country={Italy}}
\fntext[fn1]{The authors contributed equally.}
\fntext[c]{Corresponding author.}

\begin{abstract}

To enable flexible, high‑throughput automation in settings where people and robots share workspaces, collaborative robotic cells must reconcile stringent safety guarantees with the need for responsive and effective behavior. A dynamic obstacle is the stochastic, task‑dependent variability of human motion: when robots fall back on purely reactive or worst‑case envelopes, they brake unnecessarily, stall task progress, and tamper with the fluidity that true Human–Robot Interaction (HRI) demands. In recent years, learning-based human-motion prediction has rapidly advanced, although most approaches produce worst-case scenario forecasts that often do not treat prediction uncertainty in a well-structured way, resulting in over-conservative planning algorithms, limiting their flexibility. This paper introduces Uncertainty-Aware Predictive Control Barrier Functions (UA-PCBFs), a unified framework that fuses probabilistic human hand motion forecasting with the formal safety guarantees of Control Barrier Functions (CBFs). In contrast to CBFs and other variants, our framework allows for a dynamic adjustment of the safety margin thanks to the human motion uncertainty estimation provided by the deep-learning forecasting module. Thanks to the awareness of prediction uncertainty, UA-PCBFs empower collaborative robots with a deeper understanding of future human states, facilitating more fluid and intelligent interactions through informed motion planning. We validate UA-PCBFs through comprehensive real-world experiments with an increasing level of realism, including automated setups (to perform exactly repeatable motions) with a robotic hand and direct human-robot interactions (to validate promptness, usability, and human confidence). Relative to state-of-the-art HRI architectures, UA‑PCBFs show better performance in task-critical metrics, significantly reducing the number of violations of the robot's safe space during interaction with respect to the state-of-the-art. Data and code will be released upon acceptance.
\end{abstract}

\begin{keyword}


control barrier functions \sep human-robot cooperation
\sep hand trajectory forecasting \sep motion 
planning \sep collision avoidance

\end{keyword}

\end{frontmatter}




\section{Introduction}\label{sec:introduction}




Human-Robot Interaction (HRI) must adhere to stringent safety constraints to ensure reliable and efficient interaction between humans and robots. With the transition towards Industry 5.0 and the integration of collaborative robots (cobots) into human workspaces, there is the potential to increase productivity, improve worker ergonomics, and enable more flexible automation~\cite{robinson2023robotic}. However, achieving seamless and safe interaction, fundamental for the Industry 5.0 core task of Human-Robot Collaboration, remains a significant challenge due to the unpredictability of human movement. Unlike traditional robotic systems operating in structured environments with predefined tasks, mixed human-robot systems must dynamically adapt to human behavior while ensuring safety and efficiency~\cite{arents2021human}.

One of the key issues in HRI is the inherent variability and uncertainty in human motion. Human operators exhibit natural, non-deterministic behavior, which makes it difficult for robots to predict their movements accurately and therefore plan/re-plan their trajectory accordingly~\cite{sampieri2022pose,duarte2018action,li20}. This unpredictability can lead to inefficiencies, interruptions, or even safety hazards if the robot reacts too conservatively or aggressively. To manage this unpredictability, traditional control approaches typically employ predefined safety zones or reactive obstacle avoidance strategies, which limit adaptability and often result in reduced operational efficiency~\cite{lo2021towards}. 

Control Barrier Functions (CBFs) have been widely explored as a means to enforce safety constraints while maintaining real-time adaptability in robotic systems. 
CBFs are particularly well-suited for HRI due to their ability to enforce safety constraints in a minimally invasive and mathematically rigorous manner. Unlike traditional motion planning techniques, which often separate planning and control or require replanning when unexpected human motion is detected, CBFs operate directly at the control level, enabling real-time safety enforcement and offering provable guarantees on constraint satisfaction without sacrificing responsiveness or adaptability~\cite{he2021rule,palleschi2021fast,byner2019dynamic}. 

However, conventional CBFs primarily rely on instantaneous observations and are often designed for rigid-body obstacle avoidance. This leads to a purely \textit{reactive} control system, which is poorly suited for highly dynamic environments. Additionally, conventional CBF approaches fail to account for noise in human pose estimation, further complicating the task of achieving smooth and proactive interaction. Existing approaches extend CBFs to account for these shortcomings, such as Predictive Control Barrier Functions (PCBFs) \cite{breeden2022predictiveCBF}, which incorporate predictions on the evolution of the system into the CBF definition, and Stochastic Control Barrier Functions (SCBFs), which introduce stochasticity directly into the system's definition. However, neither approach fully captures the uncertainty inherent in human motion predictions: PCBFs assume deterministic forecasts, whereas SCBFs require full probabilistic modeling of system dynamics, an impractical requirement when dealing with human behavior.

\begin{figure}[t]           
  \centering
  \includegraphics[width=\linewidth,keepaspectratio]
    {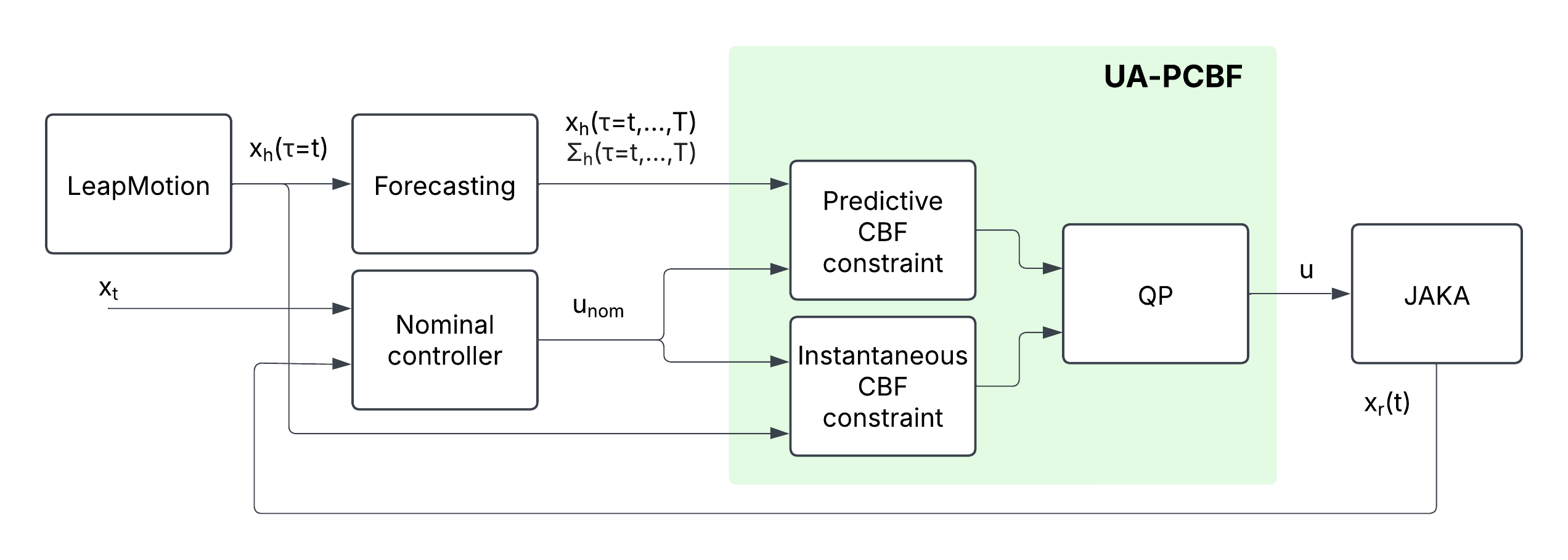}  
  \caption{Block diagram of the proposed framework. $x_t$ and $x_r(t)$ denote the target and current Cartesian position of the end effector respectively, $x_h(\tau=t)$ denotes the current Cartesian position of the hand, $x_h(\tau=t,\dots,T)$ and $\Sigma_h(\tau=t,\dots,T)$ denote the predicted sequences of cartesian positions and covariance matrices respectively. $u_{nom}$ and $u$ denote, respectively, the nominal control and the computed probabilistically safe control.}
  \label{fig:framework_schema}
\end{figure}

To address these limitations, we propose a novel framework called Uncertainty Aware-Predictive Control Barrier Functions (UA-PCBFs). This approach extends traditional CBF formulations by incorporating the human hand movement uncertainty, thanks to a deep learning forecasting model capable of 
providing an uncertainty measure for the future hand position. 
Thanks to that, our framework enables robots to proactively adjust their trajectories based on estimated future human behavior, rather than relying solely on reactive responses,  accounting also for the uncertainty of the estimation. This predictive capability enhances both safety and efficiency, allowing for more intuitive and cooperative interactions between humans and robots.
The forecasting approach is an autoregressive Deep Learning (DL) model based on LSTM~\cite{hochreiter1997long}. The non-linear nature of this model is the ability to predict the hand movement, capturing both short and long-term temporal dependencies as well as complex kinematic patterns, while at the same time can achieve an execution speed fast enough to match the robot control's requirements. Furthermore, since it is based solely on computer vision estimated 3D hand pose estimation, it guarantees fast applicability in various contexts.

To the best of our knowledge, this work is the first to integrate epistemic prediction uncertainty directly into predictive CBFs—embedding it in the safe‑set definition and enabling dynamic adjustment of the safe‑space barrier based on the current uncertainty level—without making additional assumptions about how that uncertainty will evolve. 

We validate our framework by performing real-world experiments on a \mbox{6-DoF} robot in a manufacturing facility, showing how our system can handle complex scenarios. Besides validation tests with a human operator, to guarantee reproducibility, we employ a mockup hand operated by another robot to perform the tests. Results show that our method breaks down the state-of-the-art CBFs by an order of magnitude, showing a significant increase in the safety of interactions.
\Cref{fig:framework_schema} shows a complete overview of our proposed framework.

In summary:

\begin{itemize}
\item We propose UA-PCBF, a framework that models the human dynamics using a neural network-based predictor that offers flexibility and generalization to capture complex, erratic, human behaviour without explicitly modeling it; 

\item Our UA-PCBF framework is implemented on a full 6-DoF robotic manipulator, demonstrating its feasibility and effectiveness in complex, realistic collaborative settings beyond simplified or low-dimensional scenarios;

\item We incorporate the uncertainty of human future movement to plan the robot movement, providing proactive safety in HRI scenarios;

\item Our method shows superior performance in terms of the number and gravity of robot's safe-space violations compared to state-of-the-art approaches, thereby improving both safety and efficiency in shared workspaces;

\item Our setup requires no markers or wearable devices, relying solely on vision-based 3D hand pose estimation. This makes the system practical and easily extensible to real-world industrial or collaborative applications.

\end{itemize}


The rest of the paper is structured as follows: \Cref{sec:related} briefly recalls the literature on CBFs and forecasting in HRI scenarios, motivating our work. \Cref{sec:method} presents our framework, from the 3D hand trajectory forecasting to the UA-PCBF. \Cref{sec:experiments} describes our experimental setups, and \Cref{sec:results} shows the results and discusses them. Finally, \Cref{sec:conclusions} draws some conclusions of our work and presents some final considerations.

\section{Related works}\label{sec:related}

In this section we review the state of the art that motivates the introduction of our approach. In \Cref{sec:cbf-sota} we survey Control Barrier Functions (CBFs), ranging from their classic reactive formulation and exploring their predictive, stochastic and uncertainty-aware extensions, highlighting both strengths and shortcomings in dynamic, safety-critical settings. Then, in \Cref{sec:forecasting_rel_works}, we examine human motion forecasting methods, motivating the choice of hand trajectory models as opposed to 
full-body pose predictors, discussing their real-time applicability and treatment of uncertainty. We conclude the review by identifying the unaddressed needs for a unified framework that fuses probabilistic hand‑motion prediction with formally guaranteed safety, paving the way for our UA‑PCBF formulation.

\subsection{Control Barrier Functions}
\label{sec:cbf-sota}
CBFs are widely used to ensure safety in real-time control systems by enforcing forward invariance on a predefined safe set~\cite{ames2019control}. However, traditional CBFs are inherently reactive, guaranteeing safety only in the current state without explicitly considering the future evolution of the system~\cite{cohen2020approximate-control-cbf}. This myopic behavior can lead to situations where large or rapid control actions are required, potentially resulting in unsafe interactions with humans or necessitating overly conservative safe sets that degrade task performance~\cite{breeden2023highdegreeCBF}.

To address these limitations, several extensions to the standard CBF framework have been proposed. High Order CBFs~\cite{xiao2021high-orderCBF} and Exponential CBFs~\cite{nguyen2016exponentialCBF} introduce implicit look-ahead capabilities through the design of class-$\mathcal K$ functions. While these methods offer improved performance~\cite{xiao2021high-orderCBF,nguyen2016exponentialCBF,breeden2023highdegreeCBF}, they do not explicitly model the system's future behavior and rely heavily on the careful tuning of class-$\mathcal K$ functions, which can be challenging in practice. 

A more direct approach to incorporating future system behavior is through Model Predictive Control (MPC), which has been combined with CBFs~\cite{do2024probabilisticCBF}. For instance, the Backup-CBF framework introduces backup control strategies to ensure safety when nominal plans fail~\cite{chen2021backupCBF}. However, this approach can be overly conservative and requires the explicit design of backup controllers and sets. Predictive CBFs (PCBFs)~\cite{wabersich2022predictiveCBF}, on the other hand, offer a less conservative alternative by encoding the safety of a nominal trajectory over a receding horizon directly into a CBF structure. This allows the controller to proactively adjust the trajectory before it becomes unsafe~\cite{breeden2022predictiveCBFonline}, without requiring predefined backup strategies.

While MPC-based methods provide a 
structured way to reason about future safety, they often involve solving complex optimization problems whose computational cost scales up with the length of the prediction~\cite{breeden2022predictiveCBFonline}. To mitigate this, alternative model-based approaches have been proposed. For instance, Future-Focused CBFs~\cite{black2023future-focusedCBF} aim to minimize unnecessary interventions by evaluating safety along a predicted trajectory, though they have primarily been applied in autonomous driving contexts. Environment-aware CBFs (ECBFs)~\cite{hamdipoor2023environmentCBF} and observer-based methods~\cite{quan2024observer} enhance robustness to state estimation errors, particularly in dynamic environments with moving obstacles. However, these methods typically assume knowledge of the environment 
or the statistical properties of measurement errors, which limits their applicability in HRI scenarios.

Gaussian CBFs~\cite{9888130} learn a candidate barrier function by online training a Gaussian Process (GP) on noisy safety evaluations, tightening the barrier with a confidence-bound term proportional to the posterior variance of the GP. Uncertainty-Separated CBFs~\cite{10232876} build upon this by fitting two independent GPs, one for residual dynamics and one for the barrier, mixing their posteriors into the QP constraints to guarantee robustness under state-dependent perturbations.
However, both approaches enforce safety only instantaneously, without looking ahead over a prediction horizon, making them reactive and often overly conservative in highly dynamic environments. 
Furthermore, they cannot dynamically adjust the safety margin based on prediction confidence, since disturbances are treated as unknown physical functions, rather than epistemic uncertainty. 


In contrast to prior work, our approach does not assume any specific structure or predictability of the input disturbances. This feature is essential for integrating user motion forecasting models into the method, where it is not possible to accurately model the disturbance characteristics. Building on the PCBF framework introduced in~\cite{breeden2022predictiveCBF,breeden2022predictiveCBFonline}, we propose an extension that incorporates prediction uncertainty estimation into the safety margin itself.  
To the best of authors' knowledge, this is the first application of predictive CBFs that accounts for prediction uncertainty in an epistemic fashion, without requiring assumptions about its future evolution, by integrating it into the definition of the safe set. This enables proactive safety enforcement in uncertain environments, making the approach particularly suitable for HRI applications.


\subsection{Forecasting in HRI}
\label{sec:forecasting_rel_works}

The majority of forecasting approaches in HRI focus on full-body pose prediction, where future human movements are estimated using whole-body skeletal representations~\cite{sampieri2022pose,avogaro2024exploring,guo2023back}. While this provides an overview of human motion dynamics, it often lacks the granularity required for fine-grained interactions, particularly those involving the hands.

Full-pose forecasting typically leverages standard 3D human pose estimation methods, which are well-established for capturing body motion but are often coarse and fail to include precise hand dynamics. Many pose forecasting approaches rely on datasets that do not explicitly capture hand kinematics, limiting their usefulness in scenarios requiring dexterous manipulation or hand-object interaction~\cite{amass,h36m}. Since the hands are the primary means of physical interaction in HRI tasks, ignoring them restricts the ability to model fine-grained human intent and limits the effectiveness of predictive models in collaborative tasks such as handovers, object manipulation, and shared tool use.

To incorporate hands in motion forecasting, two primary approaches can be considered: predicting the full skeletal hand along with the body pose~\cite{yan2024forecasting} or forecasting the 3D hand trajectory, which represents the spatial movement of the hand in an interaction volume. The first one often requires the full person to be visible by the system and can be computationally intensive. The latter is widely adopted, particularly in Virtual Reality (VR)~\cite{gamage2021so} and head-mounted device setups~\cite{bao2023uncertainty,cunico2023oo}. The latter are the most appropriate for real-time performance, and optimal hand observation point of view. 
However, head‑mounted cameras suffer from self‑occlusions, narrow fields of view, and the burden of wearable hardware. By contrast, our system uses a desktop‑mounted Leap Motion sensor—an infrared stereoscopic tracker that delivers high‑frequency, low‑latency 3D hand poses without any head‑worn equipment. This setup offers more consistent spatial coverage, reduces line‑of‑sight failures, and eliminates the user discomfort and calibration overhead inherent to egovision solutions.

Many trajectory forecasting approaches achieve high accuracy at the expense of heavy computational requirements, making them unsuitable for real-time applications~\cite{YuMa2020Spatio,giuliari2021transformer,MemoNet_2022_CVPR}. To bridge this gap, we develop a lightweight and real-time forecasting model that can efficiently predict hand trajectory while ensuring fluid and responsive interaction. 

\section{Methodology}\label{sec:method}

Our goal is to enforce a minimum separation between the robot and the operator in the context of interaction tasks.
We use simple geometric models to ensure fast and reliable distance computations, making the approach practical for real-time safety. More complex skeletal models could also be integrated into the framework by appropriately defining the distance function. 

Our safety objective is to maintain a minimum distance $d_{min}$ between:
\begin{itemize}
    \item a sphere of radius $r_{hand}$ enclosing the operator's hand, and
    \item a cylinder of radius $r_{cyl}$ and height $h_{cyl}$ enclosing the robot's last link and end-effector
\end{itemize}

The instantaneous distance, which will be used in the definitions of the barrier functions, is as follows:
\begin{equation}
    d(\tau, x) = \norm{\mathbf o -\mathbf c(x)} - r_{cyl}
\end{equation}
where $\mathbf o$ is the center of the sphere, $\mathbf c(x)$ is the point on the cylinder's axis that is closest to the sphere, obtained by projecting $\mathbf o$ onto the cylinder's axis, and $x$ is the state of the system, defined as the actual pose of the robot's Tool Center Point (TCP).

We provide this safety guarantee in a predictive, uncertainty-aware way. To achieve this, we first introduce a deep learning-based human hand forecasting model in Section \ref{sec:forecasting-methodology}. We then use the model to infuse a novel CBF formulation with probabilistic and predictive capabilities in Section \ref{sec:cbf}.

The full diagram block of our framework, with the role of each component, is shown in \Cref{fig:framework_schema}. 

\subsection{3D Hand Trajectory Forecasting}
\label{sec:forecasting-methodology}

The task of 3D hand trajectory forecasting is to predict future hand positions from past observations; here, we use the palm center as a proxy, represented in global Cartesian coordinates, since downstream modules reconstruct the full hand volume starting from the palm. During HRI tasks, robots must react within milliseconds to human unpredictable motion. 
To meet these stringent latency requirements, we designed a compact, autoregressive LSTM‐based neural model that runs in real-time (see \Cref{sec:implementation-details}) while striking an optimal trade-off between efficient forecasting and capturing the inherent unpredictability of human motion.

\subsubsection{Problem formulation}  
At each discrete time step \(t\), the palm position is denoted
\begin{equation}
    \mathbf{p}_t = (x_t, y_t, z_t) \in \mathbb{R}^3,
\end{equation}
where \((x_t,y_t,z_t)\) are the global 3D coordinates of the palm center. Given a historical window of length \(T_{\mathrm{in}}\),
\begin{equation}
    \mathcal{P}_{\mathrm{in}}
      = \{\mathbf{p}_{t-T_{\mathrm{in}}+1},\,\mathbf{p}_{t-T_{\mathrm{in}}+2},\dots,\mathbf{p}_{t}\},
\end{equation}
the objective is to forecast the next \(T_{\mathrm{out}}\) positions,
\begin{equation}
    \mathcal{P}_{\mathrm{out}}
      = \{\mathbf{p}_{t+1},\,\mathbf{p}_{t+2},\dots,\mathbf{p}_{t+T_{\mathrm{out}}}\}.
\end{equation}

\subsubsection{Model overview} 
We propose a lightweight encoder–decoder recurrent network  \(\mathcal{F}_\theta\) that outputs the predicted palm trajectory in the future.
To model the uncertainty of the hand's movement, at each future step, the model outputs each future Cartesian coordinate as a distribution with a certain mean and log-variance (the uncertainty). 
Formally,
\begin{equation}
  \mathcal{F}_\theta:
    \mathbb{R}^{B\times T_{\mathrm{in}}\times 3}
    \;\longrightarrow\;
    \Bigl(
      \underbrace{\mu}_{\in\mathbb{R}^{B\times T_{\mathrm{out}}\times 3}},\;
      \underbrace{\log\sigma^2}_{\in\mathbb{R}^{B\times T_{\mathrm{out}}\times 3}}
    \Bigr).
\end{equation}
where \(B\) is the batch size.

\subsubsection{Architecture}  
\begin{figure}
    \centering
    \includegraphics[width=\linewidth]{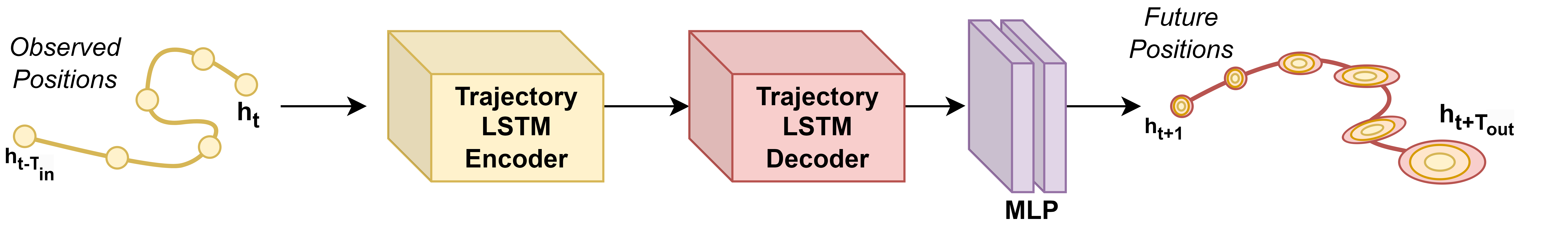}
    \caption{The architecture of the hand forecasting neural network. The model is composed of an LSTM-based encoder for spatial-temporal features and an autoregressive LSTM-based decoder that produces the final latent space states for each future prediction. Finally, the final states are used to produce the mean and log-variances of the output sequence.}
    \label{fig:hand_forecast_architecture}
\end{figure}

The model comprises three modules: (i) an \emph{encoder} that compresses the
observed 3‑D coordinates into a latent vector, (ii) an \emph{autoregressive
decoder} that unfolds this latent information step‑by‑step to produce a
sequence of latent vectors \(o_k\), and (iii) a \emph{probabilistic head} that
maps each \(o_k\) to a distribution
\([\mu_k,\,\log\sigma^2_k]\), where \(\mu_k\) is the predicted palm position
and \(\log\sigma^2_k\) quantifies its uncertainty.

The encoder is implemented as an LSTM block~\cite{hochreiter1997long}
\[
  (h^{\mathrm{enc}}_{T_{\mathrm{in}}},c^{\mathrm{enc}}_{T_{\mathrm{in}}})
    \;=\;
    \mathrm{LSTM}_{\mathrm{enc}}\bigl(\mathcal{P}_{\mathrm{in}}\bigr),
\]
where \(\mathrm{LSTM}_{\mathrm{enc}}\) has input size~3, hidden size~\(H\),
and \(L\) layers, so that
\((h^{\mathrm{enc}}_t,c^{\mathrm{enc}}_t)\in\mathbb{R}^{L\times B\times H}\) for \(t=1,\dots,T_{\mathrm{in}}\).

To generate the future trajectory latent vector, we autoregressively generate one step at a time for \(t=1,\dots,T_{\mathrm{out}}\). 
Let the encoder provide the final hidden state and denote the
last observed palm position as
\[
  (h^{\mathrm{dec}}_{0},c^{\mathrm{dec}}_{0}) \;=\;
  (h^{\mathrm{enc}}_{T_{\mathrm{in}}},c^{\mathrm{enc}}_{T_{\mathrm{in}}}),
  \qquad
  \mathbf{x}^{\mathrm{dec}}_{0} \;=\; \mathbf{p}_{t}.
\]

For each prediction step \(k = 1,\dots,T_{\mathrm{out}}\) we iterate
\begin{equation}\label{eq:nn_decoder}
\bigl(o_k,h^{\mathrm{dec}}_k,c^{\mathrm{dec}}_k\bigr) = \mathrm{LSTM}_{\mathrm{dec}}\!\bigl(
            \mathbf{x}^{\mathrm{dec}}_{k-1},\,
            (h^{\mathrm{dec}}_{k-1},c^{\mathrm{dec}}_{k-1})
        \bigr),
\end{equation}

\begin{equation}\label{eq:nn_prob}
[\mu_k,\log\sigma^2_k] =
    W_o\,o_k + b_o,
    \quad o_k\in\mathbb{R}^{B\times1\times H}, 
\end{equation}
\begin{equation}
  \mathbf{x}^{\mathrm{dec}}_{k} =
    \begin{cases}
      \mathbf{p}_{t} & \text{if k=0 (\textit{i.e.}, the first iteration)},\\[4pt]
      \mu_{k-1}            & \text{otherwise},
    \end{cases}
\end{equation}
with \cref{eq:nn_prob} being the probabilistic head based on a multi-perceptron layer with $W_o$ weights and $b_o$ bias.

Finally, stacking over \(t=1,\dots,T_{\mathrm{out}}\) yields the output $[\mu, \log\sigma^2]$, with \(\mu\in\mathbb{R}^{B\times T_{\mathrm{out}}\times 3}\) and \(\log\sigma^2\in\mathbb{R}^{B\times T_{\mathrm{out}}\times 3}.\)


\subsubsection{Training objective}  
Denote the ground‑truth future sequence by \(\mathbf{p}^{\mathrm{true}} \in \mathcal{P}_{\mathrm{out}}\). We minimize a composite loss of Gaussian negative log-likehood (NLL) and mean squared error (MSE). 
The NLL is defined as 
\begin{equation}
  \mathcal{L}_{\mathrm{NLL}}
    = \frac{1}{B}\sum_{i=1}^B\sum_{t=1}^{T_{\mathrm{out}}}
       \biggl[\tfrac12\log\sigma^2_{i,t}
             + \frac{(\mathbf{p}^{\mathrm{true}}_{i,t}-\mu_{i,t})^2}{2\,\sigma^2_{i,t}}
       \biggr],
\end{equation} and the MSE is defined as 
\begin{equation}
  \mathcal{L}_{\mathrm{MSE}}
    = \frac{1}{B\,T_{\mathrm{out}}}\sum_{i=1}^B\sum_{t=1}^{T_{\mathrm{out}}}
       \|\mu_{i,t}-\mathbf{p}^{\mathrm{true}}_{i,t}\|_2^2.
\end{equation}

The final loss is $
  \mathcal{L} = \rho\,\mathcal{L}_{\mathrm{NLL}} \;+\;\omega\,\mathcal{L}_{\mathrm{MSE}}
$,
with hyperparameters \(\rho,\omega>0\). 

The Gaussian NLL term not only encourages accurate mean predictions but also explicitly penalizes mis‐estimated variances, thereby guaranteeing non‐degenerate uncertainty estimates. Concretely, the NLL contains a \(\tfrac12\log\sigma^2\) penalty which diverges if \(\sigma^2\to0\), and a residual term \(\tfrac{(\mathbf{p}^{\mathrm{true}}-\mu)^2}{2\,\sigma^2}\) which blows up if the variance is underestimated when the mean prediction errs.  This creates an inherent trade‑off: the model may only predict very low variance when it is confident that its mean is highly accurate, else the loss becomes prohibitively large.  As a result, the network is driven to learn meaningful, data‑dependent uncertainty rather than collapsing to a trivial ``mean with zero variance.''

\subsection{Control Barrier Functions}
\label{sec:cbf}

We consider the control‐affine system
\begin{equation}
  \dot x = f(x) + g(x)\,u
\end{equation}
and define a safe set 
$\mathcal S = \{\,x : h(x)\le0\}$ via a continuously differentiable barrier function $h$. Given a nominal control signal $u_{nom}$, we compute the control signal $u^\star$ that best approximates it while ensuring that the state of the system never leaves the safe set, by solving the following Quadratic Program (QP)
\begin{equation}
    \begin{split}
        u^\star = \underset{u\in\mathbb R^m}{\arg\min}&\;\;\;\frac{1}{2}\norm{u-u_{nom}}^2\\
        s.t.&\;\;\;L_gh\,u\leq -L_fh - \alpha(h)
    \end{split}
\end{equation}
where $L_fh$ and $L_gh$ are the Lie derivatives of $h$ along $f$ and $g$, respectively, and $\alpha(h)$ is a class-$\mathcal K$ function. When $h(x)$ is well below zero, the constraint is inactive and $u^\star\approx u_{nom}$.  As $h(x)$ approaches the boundary, the QP smoothly adapts $u^\star$ to maintain safety.

While effective, this approach is myopic since it enforces the safety constraints only at the current state, which can lead to over-conservatism or infeasibility in constrained scenarios.

\subsection{Predictive Control Barrier Functions}
\label{sec:pcbf}

To address the limitations of standard CBFs, Predictive CBFs (PCBFs) introduce a receding-horizon safety metric, evaluating safety along a predicted nominal trajectory over a finite time horizon $[t, t+T_{out}]$.

Given a nominal control policy $u_{nom}(t,x)$, the evolution of the system over a future horizon $T_{out}$ is defined as
\begin{equation}
    \frac{d}{d\tau}p(\tau,t,x)=f(p(\tau,t,x)) + g(p(\tau,t,x))u_{nom}
\end{equation}
where $p(\tau,t,x)$ is a path function representing the state of the system at time $\tau$ assuming the system is driven by $u_{nom}$.

To assess whether the predicted evolution of the system is safe or not, the 
barrier function is evaluated over the entire trajectory in the $[t,t+T_{out}]$ time horizon, resulting in a Predictive Control Barrier Function (PCBF)
\begin{equation}
    h_p(\tau,t,x) = h(\tau,p(\tau,t,x)) - m(R(\tau,t,x) - t)
\end{equation}
where $h(\tau,p(\tau,t,x))$ is the function that measures safety along the path, $R(\tau,t,x)$ is the earliest time on the horizon where a safety violation might occur, and $m$ is a class-$\mathcal K$ margin function.

The control signal that adjusts the evolution of the system is then obtained by solving the following constrained QP
\begin{equation}
    \begin{split}
        u^{\star}_{pcbf} = \underset{u\in\mathbb R^m}{\arg\min}&\;\;\;\frac{1}{2}\norm{u-u_{nom}}^2\;\;\;\;\;\;\; \forall \tau\in[t,t+T_{out}]\\
        s.t.&\;\;\;L_gh_p\,u\leq -L_fh_p - \alpha(h_p).
    \end{split}
\end{equation}

PCBFs improve upon CBFs by adopting a proactive strategy that considers the predicted evolution of the system, enabling early corrective actions and reducing overly conservative behaviors. However, PCBFs assume that the predicted trajectory is deterministic and accurate, disregarding the uncertainty inherent in real-world forecasting models, especially in highly dynamic environments that are commonly encountered in HRI.

\subsection{Uncertainty Aware-Predictive Control Barrier Functions}

To enforce both reactive and predictive safety constraints under uncertainty, we now introduce our Uncertainty Aware-Predictive Control Barrier Functions (UA-PCBFs). In UA-PCBFs, the uncertainty is used to dynamically scale a predictive barrier function, relaxing and tightening safety requirements based on the confidence of the prediction.
We then combine a reactive and a predictive constraint through the introduction of slack variables. Embedding these uncertainty-aware constraints into a single QP allows us to obtain a controller that smoothly balances responsiveness and safety compliance.

\subsubsection{Treating forecast uncertainty}

To introduce uncertainty into the barrier function formulation, we first recover the covariance matrix from the forecaster's log-var output at time $\tau$
\begin{equation}
    \sigma^2(\tau) = \exp\left(\frac{1}{2}\log\sigma^2(\tau)\right)\in\mathbb R^3\implies \Sigma(\tau)=\text{diag}(\sigma^2(\tau))\in\mathbb R^{3\times3}.
\end{equation}

Since we are dealing with interaction tasks we focus on the direction of interaction, which is identified by the unit vector $u$ that goes from the cylinder to the center of the hand, along the shortest path. By projecting the covariance matrix $\Sigma(\tau)$ along $u$, we get a scalar measure of dispersion that the covariance matrix induces on the interaction axis
\begin{equation}
    \sigma_{proj}(\tau, x)=u^T\Sigma(\tau)u,\;\;\;\;u=\frac{\mathbf o - \mathbf c(x)}{\norm{\mathbf o - \mathbf c(x)}}
\end{equation}

To handle the cases in which the forecaster overestimates the log-variance, we clamp the obtained projected variance to $d_{min}$, ensuring that uncertainty is never overestimated at the control phase
\begin{equation}
    \overline\sigma(\tau, x) = \begin{cases}0 & \tau = 0\\
        \gamma\sigma_{proj}(\tau, x) & \gamma\sigma_{proj}(\tau, x)<d_{min}\\
        d_{min} & \text{otherwise},
    \end{cases}
\end{equation}
where $\gamma$ is a parameter that allows for the tuning of the influence of the uncertainty on the controller. The uncertainty at time $\tau=0$ is set to zero by definition, since the first point in the future trajectory will always be the current position of the operator's hand, and thus it is not affected by prediction accuracy.

\subsubsection{Uncertainty-aware barrier function}
\label{sec:uapcbf}
We can now define the probabilistic barrier function $h_{ua}$
\begin{equation}
    h_{ua}(\tau, x) = d_{min} + \overline\sigma(\tau, x)- d(\tau, x)\;\;\;\tau\in [t,t+T_{out}]
\end{equation}

The introduction of the uncertainty $\overline\sigma$ results in a dynamic modulation of the safety distance between robot and operator. When the forecasting is uncertain about its prediction, the barrier function adapts itself by ensuring a higher safety distance is maintained, and vice versa.

To ensure rapid reaction time, we mix an instantaneous barrier function, based on the current hand position, and a predictive barrier function. We achieve this by introducing slack variables $\delta_r$ and $\delta_p$ in the two constraints we impose on the QP
\begin{align}
\label{eq:uapcbf_constraints}
\begin{split}
L_gh_{ua}(0, x)\,u&\leq -L_fh_{ua}(0, x) - \alpha(h_{ua}(0, x)) + \delta_r,\;\;\;\delta_r\ge 0\\
L_gh_{ua}(\tau, x)\,u&\leq -L_fh_{ua}(\tau, x) - \alpha(h_{ua}(\tau, x)) + \delta_p,\;\;\;\delta_p\ge 0.
\end{split}
\end{align}
The slack variables $\delta_r$ and $\delta_p$ affect the reactive (instantaneous) and predictive barrier functions, respectively. They are nonnegative quantities, introduced to ensure the feasibility of the problem by allowing minimal, well-controlled violations when either constraint becomes too restrictive.

The control input is then computed by solving the QP
\begin{equation}
    \begin{split}
        u^{\star}_{ours} = \underset{u\in\mathbb R^m}{\arg\min}&\;\;\;\frac{1}{2}\norm{u-u_{nom}}^2+\lambda_{r}\delta_{r}^2+\lambda_{p}\delta_{p}^2\\
        s.t.&\;\;\text{eqn. (\ref{eq:uapcbf_constraints})}
    \end{split}
\end{equation}
where $\lambda_{r}$ and $\lambda_{p}$ are penalty parameters that discourage the violation, respectively of the reactive and predictive constraints, unless strictly necessary. As such, the slack variables in the QP act as decision variables, allowing the controller to dynamically relax the safety constraints based on current task conditions and predicted future risk. We set the penalty terms as 
\begin{equation}
\label{eq:lambda_p}
    \lambda_r = 100,\;\;\;\;\lambda_p=\lambda_r - \frac{\gamma\overline\sigma}{d_{min}}.
\end{equation}
The penalty for the reactive term, $\lambda_r$, is fixed to always ensure that the optimizer avoids relaxing the constraint related to the current position of the hand. The penalty for the predictive term $\lambda_p$ is influenced instead by the uncertainty in the prediction itself. When the uncertainty is low $\lambda_p\approx\lambda_r$ so the optimizer is discouraged from violating the predictive constraint. In contrast, when the uncertainty is high $\lambda_p\approx\lambda_r-\gamma$ so the optimizer is allowed to violate the constraint in a controlled manner. In summary, the uncertainty coming from the prediction model is integrated in two ways: directly into the barrier function, by dynamically modulating the safety margin, and indirectly inside the QP, allowing the controller itself to relax or tighten the predictive constraint when needed.

\section{Experimental setup}\label{sec:experiments}

This section details the experimental protocols that we used to validate both the hand-forecasting model and the proposed UA-PCBF approach, in progressively more realistic settings. In Sections \ref{sec:leap-data}--\ref{sec:forecasting_system}, we describe the data collection setup, which uses a Leap Motion v2 sensor to record 3D hand trajectories under different motion patterns. These recordings are structured into a dataset that is then used in training and testing the deep learning-based forecasting model. We present the model's architecture and the related evaluation metrics. Finally, in \Cref{sec:experimental_protocol} we present the two real-world scenarios in which the proposed UA-PCBF framework was evaluated: a highly repeatable mockup experiment using a hand model actuated by a 6 DoF manipulator, and a human-in-the-loop handover task assessing performance in realistic HRI conditions. 
\Cref{fig:robotic_cell_with_axes} shows the robotic cell where the real-world experiments are performed. In the \Cref{fig:robotic_cell_with_axes} (a), the reference frames of each experiment unit: the robot's base and TCP, and the Leap Motion, can be seen. In \Cref{fig:robotic_cell_with_axes} (b), the bounding regions for both robot and operator are shown. 

For each scenario, we define quantitative metrics spanning spatial accuracy, motion efficiency, and safety compliance. We conclude the section with implementation specifics on both the forecasting network and the UA-PCBF robot controller implementation.

\subsection{Hand Tracking Data}\label{sec:leap-data}

To track the hand position, we rely on the Leap Motion v2 device. This device can track the hands' tracking data in real-time (up to 120 Hz). 
We leverage this tool to obtain the motion trajectory of the hand, both in real-world experiments and for data collection for the training data.
Using the Leap Motion v2, we created a dataset consisting of several trajectories of human hand motion and orientation, for a total of 160k individual samples, collecting over 33k subsequences of 1 second each of hand positions at 30 Hz. 
Considering all the subsequences, we have a per‐sequence average speed of $0.1664 \pm 0.1210\ \mathrm{m/s}$ with a maximum observed speed of $7.4130\ \mathrm{m/s}$, and a per‐sequence average acceleration of $1.4448 \pm 1.0008\ \mathrm{m/s}^2$, with a maximum observed acceleration of $256.4722\ \mathrm{m/s}^2.$
The data will be released with the code upon acceptance for applicability.

\subsection{Forecasting System}\label{sec:forecasting_system}

The forecasting model, detailed in \Cref{sec:forecasting-methodology}) has been trained on the dataset collected using the Leap Motion v2 device (see \Cref{sec:leap-data}). The next sections detail how the forecasting system is evaluated against different time horizons and which metrics are used to evaluate it. 

To assess the forecasting performance, we trained the model on various time horizons, and we opted for 1000 ms for the real-world experiments, as is typical in human-robot interaction and collaboration tasks~\cite{sampieri2022pose}.

Trajectory forecasting is measured in terms of final displacement error (FDE) and average displacement error (ADE), standard metrics in trajectory forecasting~\cite{mohamed2020social}. These metrics quantify how well the predicted hand trajectory aligns with the ground truth over a given prediction horizon.

The ADE measures the average Euclidean distance between the predicted and ground truth trajectory points over all time steps
\begin{equation}
    ADE = \frac{1}{T_{\text{out}}} \sum_{t=1}^{T_{\text{out}}} \|\mathbf{h}_t^{\text{pred}} - \mathbf{h}_t^{\text{true}}\|_2,
\end{equation}
where \( T_{\text{out}} \) is the number of future time steps, and \( \mathbf{h}_t^{\text{pred}}, \mathbf{h}_t^{\text{true}} \in \mathbb{R}^3 \) are the predicted and ground truth hand positions at time \( t \).

The FDE evaluates the positional error at the final predicted time step
\begin{equation}
    FDE = \|\mathbf{h}_{T_{\text{out}}}^{\text{pred}} - \mathbf{h}_{T_{\text{out}}}^{\text{true}}\|_2.
\end{equation}



The model can run at 30 Hz, processing the Leap Motion data with a sliding window of the last $N$ tracking events and producing the subsequent sequence of $M$ positions and rotations, as explained in \Cref{sec:forecasting-methodology}

\subsection{Implementation details}\label{sec:implementation-details}
This section presents implementation details for the two main software components: the 3D hand forecasting and the robotic control. 

\subsubsection{3D Hand Forecasting}
The 3D hand forecasting model has been trained with Python version 3.11 and CUDA 12.1 on an Ubuntu 22.04 operating system and an RTX 3090. The training lasted for 200 epochs using the AdamW optimizer with an initial learning rate of 1e-4, with a cosine annealing scheduler. The model run at 30 Hz on an RTX 3060 in inference.

\subsubsection{Robot and Control}

The experiments were carried out with a JAKA ZU 5 6-DoF robotic manipulator, controlled through a dedicated workstation. Figure \ref{fig:robotic_cell_with_axes} shows the robot placed in its collaborative cell, together with the relevant reference frames. 


To control the robot, a ROS2 Jazzy environment was set up on the workstation. Dedicated ROS2 nodes were then wrote to encapsulate JAKA's Python SDK v.2.2.2, which exposes API functions related to robot status, control and communication. The only exception were the functions related to kinematics (i.e. direct/inverse kinematics and Jacobian computation), of which the SDK's implementations were deemed either too slow, in the case of direct/inverse kinematics, or non-existent, in the case of Jacobian computation. For those functions, the Robotics Toolbox for Python \cite{rtb} was used. 

The robot was controlled with a task space proportional velocity controller. Given the current joint angles $q$, the end‐effector pose $(p_{\mathrm{c}},R_{\mathrm{c}})$ is obtained by forward kinematics.  
 A proportional feedback law then computes the desired twist $v_{\mathrm{ref}}\in\mathbb{R}^6$ based on position and orientation errors
 \begin{equation}
 e_p = p_{\mathrm{d}} - p_{\mathrm{c}}, 
 \quad
 e_r = \tfrac12\bigl(R_{\mathrm{d}}R_{\mathrm{c}}^{T} - R_{\mathrm{c}}R_{\mathrm{d}}^{T}\bigr)^\vee,
 \end{equation}
 where $(\cdot)^\vee$ extracts the axis–angle vector.  The task space twist is then computed with the proportional law
 \begin{equation}
     v_{\mathrm{ref}} = (K_Pe_p, K_re_r)
 \end{equation}


Finally, the twist is mapped into the joint space using the pseudo-inverse of the robot's Jacobian 
\begin{equation}
     \dot q_{ref} = u_{nom} = J(q)^{+}v_{\mathrm{ref}}
\end{equation}

\subsection{Experimental protocol}
\label{sec:experimental_protocol}

\begin{figure}[h]          
  \centering
  \includegraphics[width=\linewidth,keepaspectratio]
    {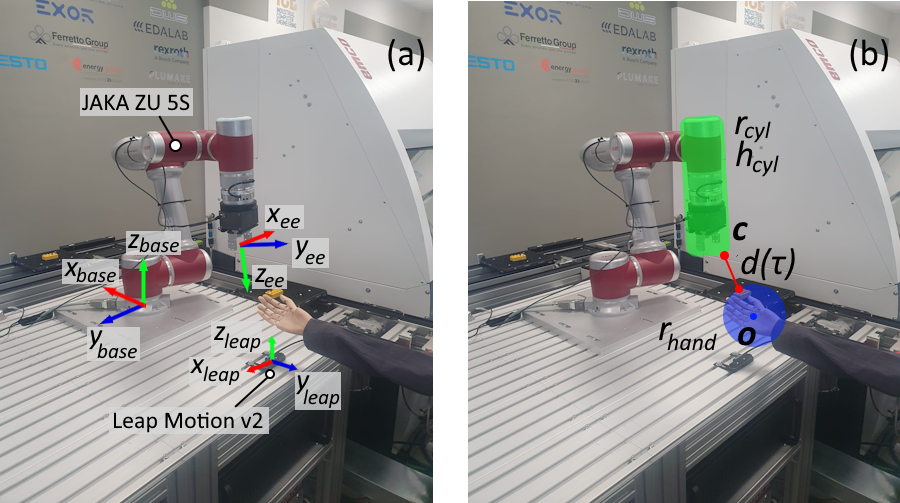}  
  \caption{The experimental setup for the hand mockup experiment. (a) shows the relevant frames of reference. (b) shows the bounding regions of both robot and hand model, with the corresponding distance $d(\tau)$.}
  \label{fig:robotic_cell_mock}
\end{figure}

To evaluate the performance of the proposed framework, two experimental scenarios were designed. In the first one, a hand model was used to collect consistent, repeatable results in simplified conditions. In the second scenario, the system was tested by a human operator, showcasing its applicability in realistic conditions.

\subsubsection{Hand mockup experiment}

\begin{table}[ht]
  \centering
  \begin{threeparttable}
    \caption{Summary of the parameters used during the runs of the mock hand experimental scenario.}
    \label{tab:cbf_params}
    \scriptsize
    \begin{tabular}{ccccc}\toprule
      \textbf{Method} & \textbf{$\alpha$(h)} & \textbf{m($\tau$)}
        & \textbf{$\gamma$} & \textbf{$\lambda_p$} \\\midrule
      CBF     & 125h     & –      & –     & –         \\
      PCBF    & 125h     & $\tau^2$    & –     & –         \\
      UA-PCBF & 125h     & $\tau^2$    & 0     & *         \\
      UA-PCBF & 125h     & $\tau^2$    & [0,0.5,1,2.5,5] & $\lambda_r$   \\
      UA-PCBF & 125h     & $\tau^2$    & [0,0.5,1,2.5,5] & *         \\ 
      \bottomrule
    \end{tabular}
    \begin{tablenotes}\footnotesize
      \item[*] computed with equation \ref{eq:lambda_p}.
    \end{tablenotes}
  \end{threeparttable}
\end{table}



A wooden human hand model was rigidly mounted to the end-effector of a UR5e robotic manipulator to ensure highly repeatable and controlled hand trajectories across experimental trials. The setup is shown in Figure \ref{fig:robotic_cell_mock}. During each run, the hand executed a sequence of rapid upward motions, characterized by a peak velocity of $1.0~\mathrm{m/s}$ and an acceleration of $3.5~\mathrm{m/s^2}$. These motion parameters were selected to emulate sudden, high-dynamic hand movements that are particularly challenging for real-time obstacle avoidance algorithms.

Simultaneously, a JAKA robotic arm (i.e., the cobot in our HRI cell) performed a linear sweeping motion above the hand, moving back and forth along a path orthogonal to the hand's trajectory. This setup was designed to simulate a dynamic, shared workspace scenario in which the robot must continuously adapt its motion in response to fast, human-like movements.

Five experimental trials were conducted, each employing a different control strategy: two baseline methods (CBF and PCBF), two ablated variants of the proposed method (with $\gamma = 0$ and $\lambda_p = \lambda_r$), and the proposed UA-PCBF approach. The control parameters used in each configuration are summarized in Table~\ref{tab:cbf_params}. For each trial, five independent runs were recorded to enable statistical analysis.

\subsubsection{Human operator experiment}

\begin{figure}[ht]           
  \centering
  \includegraphics[width=\linewidth,keepaspectratio]
    {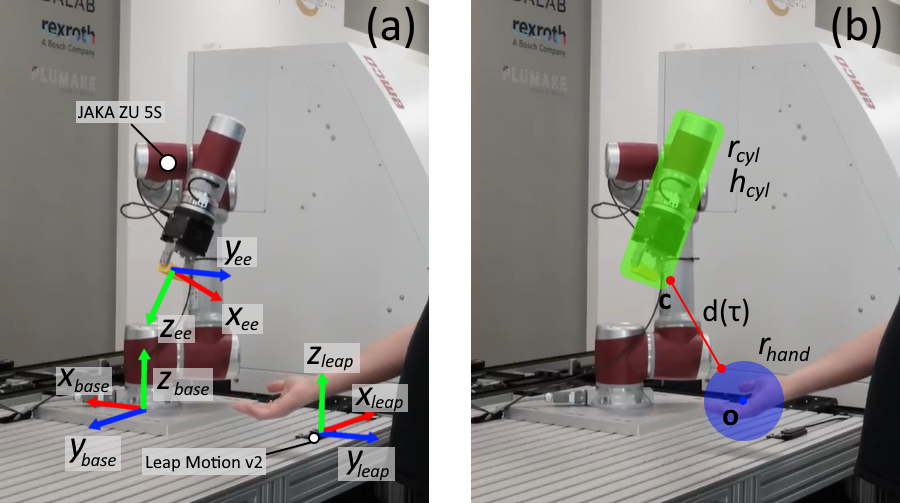}  
  \caption{The experimental setup for the human operator experiment. (a) shows the relevant frames of reference. (b) shows the bounding regions of both robot and operator hand, with the corresponding distance $d(\tau)$.}
  \label{fig:robotic_cell_with_axes}
\end{figure}

\begin{figure}[t]
    \centering
    \includegraphics[width=\linewidth]{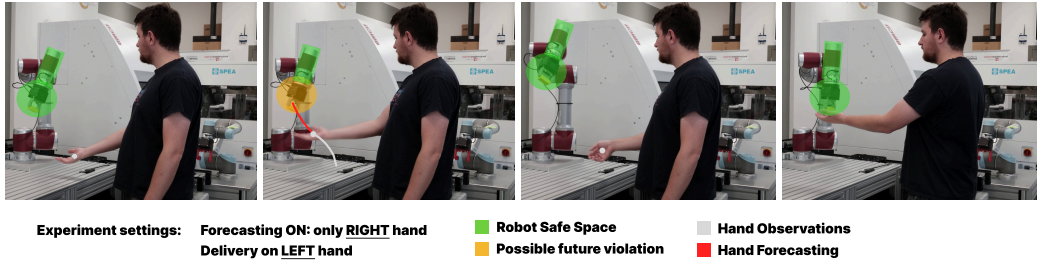}
    \caption{Example of the framework in a real-world implementation. In this example, the right hand is the only one tracked with the forecasting method. Note that this is a testing choice, not a limitation of the framework. The operator stands in front of the robot, and the hand is tracked in 3D using a Leap Motion V2 device. The device streams the data in real-time to the neural network that predicts the future position of the hand with a 1000ms time horizon. When the operator moves the robot control, seeing that the hand will collide within the future time horizon, adjusts the planning by keeping a safe distance from the operator's hand. Finally, the delivery is performed on the left hand.}
    \label{fig:real-world}
\end{figure}

To evaluate the applicability of the method in realistic HRI conditions, a handover task was designed. The robot carried out a pick-and-place operation, picking up an object from a bay and placing it on the operator's left hand. During the motion between the pick and place operations, the operator was asked to replicate the movement carried out by the UR5e manipulator in the hand mockup experiment. 
\Cref{fig:robotic_cell_with_axes} shows the reference frames of the different systems involved in the experiment and the bounding regions, while in \Cref{fig:real-world} we show the execution of one take of the experiments, with four frames of the task. We show that the right hand is tracked and forecasted, while our method uses the future prediction and its uncertainty to adjust the robot's path planning.

In this scenario, three runs were collected, using the parameters that showed the best performance with respect to interaction metrics, which will be discussed in the following section.

\subsubsection{Metrics}
To evaluate the performance of the proposed framework in handling human motion, executing collaborative tasks, and ensuring safety, we consider six interaction-based performance metrics as proposed in \cite{tuli2022humanquality,jost2020human-robot-interaction}. These metrics are grouped into three categories—spatial accuracy, dynamics, and safety compliance—to provide a comprehensive assessment of the human–robot interaction.


\begin{enumerate}
    \item Spatial accuracy 
    \begin{itemize}
        \item Mean hand-TCP distance (meters): the average Euclidean distance between the palm of the human hand and the robot's Tool Center Point (TCP) over the course of the execution.
        \item Total TCP path length (meters): the cumulative length of the trajectory traversed by the TCP during the task. 
    \end{itemize}
    \item Dynamics 
    \begin{itemize}
        \item Average TCP velocity (meters per second): the mean of the TCP's instantaneous linear speeds during execution.
        \item Total execution time (seconds): The overall duration required to complete the task from start to finish.
    \end{itemize}
    \item Safety compliance 
    \begin{itemize}
        \item Number of safety limit violations: the total number of instances in which the safety function $h$ exceeds a predefined threshold, indicating a breach of the robot’s safety boundary. In our evaluation, this threshold is set to 10\,mm.
        \item Mean violation magnitude (meters): how far the TCP breached the safety boundary, when a violation occurred, averaged over all breach events.
    \end{itemize}
\end{enumerate}



\section{Experimental results}\label{sec:results}

In this section, we provide our empirical findings.  
In Section \ref{sec:forecasting_results}, we compare the hand-trajectory predictor against baselines with respect to average/end-point errors, reporting quantitative forecasting performance. In Sections \ref{sec:mock_results}-\ref{sec:operator_results}, we analyze the UA-PCBF controller's behavior in both mockup and human-in-the-loop trials, focusing on interaction accuracy, smoothness, and safety-compliance.

\subsection{Forecasting results}
\label{sec:forecasting_results}

\Cref{tab:forecasting-traditional} presents the performance of various standard methods to deal with the forecasting problem. All these approaches perform forecasting of $T_{out}$ steps from the position at time $t$, given the past observation of $T_{in}$ steps. We analyze three alternatives: a simple linear interpolation model, a particle filtering method, and a Kalman filter approach. All of them produce high errors, in particular regarding the reference time horizon of 1000 ms, for which the FDE value, \textit{i.e.}, the final position, has more than 0.18~m of error on average. 
Instead, our neural approach is more robust and achieves remarkably lower ADE and FDE values, as reported in \Cref{tab:forecasting-results}, where the metrics on different time horizons are presented. Notably, the 1000 ms reference horizon has an FDE of just 0.073~m on average, considerably lower than traditional approaches.

\begin{table}[hb]
    \centering
    \caption{Comparison between different types of forecasting methods. Metrics are presented as mean$\pm$standard deviation ($\mu\pm\sigma$) in meters (the lower the better). Leading zero omitted for formatting reasons.}
\scriptsize
\begin{tabular}{lrrrrrr}
\toprule
\textbf{Horizon (ms)} & \multicolumn{2}{c}{\textbf{Linear Interpolation}} & \multicolumn{2}{c}{\textbf{Kalman Filter}} & \multicolumn{2}{c}{\textbf{Particle Filtering}} \\
 & \textbf{ADE~$\downarrow$} & \textbf{FDE~$\downarrow$} & \textbf{ADE~$\downarrow$} & \textbf{FDE~$\downarrow$} & \textbf{ADE~$\downarrow$} & \textbf{FDE~$\downarrow$} \\

\midrule
  100   & $.004 \pm .009$ & $.007 \pm .015$ & $.038 \pm .045$ & $.044 \pm .051$ & $.007 \pm .009$ & $.009 \pm .014$ \\
  200   & $.009 \pm .014$ & $.018 \pm .028$ & $.046 \pm .052$ & $.062 \pm .069$ & $.011 \pm .013$ & $.020 \pm .028$ \\
  300   & $.016 \pm .030$ & $.036 \pm .061$ & $.058 \pm .061$ & $.084 \pm .088$ & $.018 \pm .029$ & $.037 \pm .061$ \\
  400   & $.025 \pm .034$ & $.056 \pm .077$ & $.068 \pm .071$ & $.107 \pm .110$ & $.026 \pm .034$ & $.057 \pm .076$ \\
  500   & $.033 \pm .046$ & $.076 \pm .101$ & $.078 \pm .081$ & $.127 \pm .129$ & $.035 \pm .045$ & $.077 \pm .100$ \\
  566   & $.037 \pm .050$ & $.085 \pm .111$ & $.082 \pm .083$ & $.134 \pm .132$ & $.039 \pm .049$ & $.086 \pm .110$ \\
  1000  & $.084 \pm .101$ & $.189 \pm .218$ & $.129 \pm .121$ & $.230 \pm .212$ & $.085 \pm .101$ & $.189 \pm .218$ \\
  1200  & $.109 \pm .134$ & $.238 \pm .287$ & $.153 \pm .137$ & $.270 \pm .242$ & $.110 \pm .134$ & $.239 \pm .286$ \\
\bottomrule
\end{tabular}
    
    \label{tab:forecasting-traditional}
\end{table}

\begin{table}[ht]
  \centering
  \caption{Results of the forecasting model on different time horizons (in milliseconds). Metrics are presented as mean\,$\pm$\,standard deviation ($\mu\pm\sigma$) in meters (lower is better).}
  \label{tab:forecasting-results}
  \scriptsize
  \begin{tabular}{lrr}
    \toprule
    \textbf{Horizon (ms)}
      & \textbf{ADE~$\downarrow$}
      & \textbf{FDE~$\downarrow$} \\
    \midrule
    100   & $0.003 \pm 0.003$ & $0.005 \pm 0.006$ \\
    200   & $0.006 \pm 0.008$ & $0.012 \pm 0.016$ \\
    300   & $0.009 \pm 0.012$ & $0.019 \pm 0.027$ \\
    400   & $0.013 \pm 0.016$ & $0.028 \pm 0.038$ \\
    500   & $0.016 \pm 0.021$ & $0.035 \pm 0.048$ \\
    566   & $0.018 \pm 0.024$ & $0.040 \pm 0.054$ \\
    1000  & $0.034 \pm 0.043$ & $0.073 \pm 0.086$ \\
    1200  & $0.040 \pm 0.048$ & $0.084 \pm 0.093$ \\
    \bottomrule
  \end{tabular}
\end{table}

To assess model capability of estimating the uncertainty of the predicted trajectories, we examinate the ``calibration'', following the procedure described in ~\cite{lakshminarayanan2017simple}, \textit{i.e.}, we compute for each confidence level (0.90, 0.95, 0.99), the empirical coverage defined as the fraction of the true future points that lie within the corresponding predicted confidence intervals, and then compare these empirical frequencies to their nominal values.

Our results at the 1000 ms horizon are highly encouraging: we observe empirical coverages of 90.7\%, 94.3\%, and 97.6\% for the 90\%, 95\%, and 99\% intervals, respectively. The 90\% and 95\% intervals nearly coincide with their targets, demonstrating that the model’s variance predictions are well tuned to the typical error distribution. Even the 99\% interval, despite the greater challenge of capturing rare, large deviations, exhibits only modest under‑coverage, indicating that extreme‐tail uncertainty is still largely captured.

\begin{figure}
    \centering
    \includegraphics[width=\linewidth]{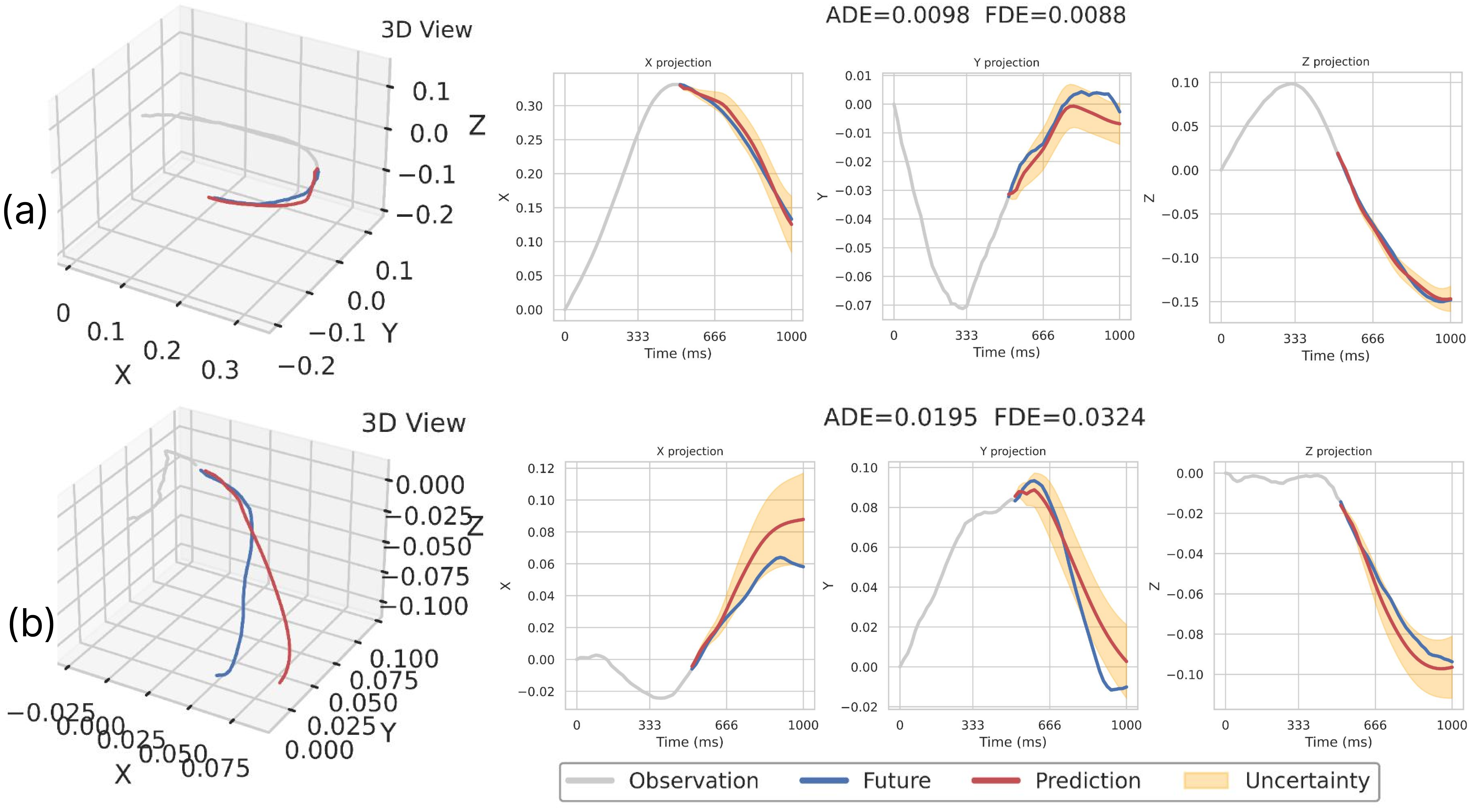}
    \caption{Qualitative examples of the forecasting model on the data validation set, with uncertainty visualization. Figure (a) shows the model that is perfectly able to follow a curved motion, with low uncertainty in X and Z axes, and a slight uncertainty (less than 2 cm) in the Y axis. Figure (b) shows the model uncertainty power: despite producing a trajectory, the displacement at the final element is high with respect to the ground truth, reflected as well in the ADE and FDE metrics. Nonetheless, the model understands that there could be some error, and the uncertainty is increased significantly towards the end of the trajectory, particularly in the X and Z axes.}
    \label{fig:forecasting-qualitative}
\end{figure}

Furthermore, we show qualitative results of the uncertainty estimation in \Cref{fig:forecasting-qualitative}. In \Cref{fig:forecasting-qualitative}~(a), an example of a trajectory with low uncertainty is shown. As can be seen, the model is perfectly able to reconstruct the circular motion with a very low ADE and FDE, and the uncertainty is overall small in value, with the only exception of the Y axis, for which, however, it is between 1-2 cm of uncertainty, therefore quite precise. In \Cref{fig:forecasting-qualitative}~(b), we can see a prediction with high ADE and FDE. In this case, the final predicted trajectory is not so close to the real one, and indeed, even if the model produces a trajectory, the uncertainty is considerably higher, denoting the uncertainty awareness intrinsic in the model. A trajectory like the one presented in \Cref{fig:forecasting-qualitative}~(b) is the perfect example of why we need a planning framework that takes into account the uncertainty of the hard-to-predict human movement.

\subsection{Hand mockup experiment}
\label{sec:mock_results}

We describe in this section the results of the hand mockup experiment, first considering the influence of the uncertainty weight parameter $\gamma$ and subsequently presenting the comparison with the baseline methods and ablated variants of the proposed method.


\begin{table}[hb]
\centering
\caption{Average values of the metrics considered obtained by the UA-PCBF method as the $\gamma$ parameter varies.  Metrics are presented as mean\,$\pm$\,standard deviation ($\mu\pm\sigma$), where values in bold represent the best results.}
\label{tab:gamma}
\scriptsize
\begin{tabular}{lllllll}\toprule
      & \textbf{Hand-TCP }            & \textbf{Trajectory }          & \textbf{Trajectory}           & \textbf{Completion  }       & \textbf{Violation}          & \textbf{Violation }              \\
$\gamma$ & \textbf{Distance}             & \textbf{Length  }             & \textbf{Velocity }            & \textbf{Time}               & \textbf{Count}              & \textbf{Magnitude}               \\
\midrule
0     & \textbf{0.32 ± 0.01} & 2.17 ± 0.16          & 0.33 ± 0.02          & 6.47 ± 0.01        & 8.4  ± 2.1         & 0.024 ± 0.005           \\
0.5   & \textbf{0.32 ± 0.01} & 2.07 ± 0.12          & 0.32 ± 0.01          & 6.42 ± 0.02        & 5.2 ± 4.8          & 0.016 ± 0.1             \\
1     & \textbf{0.32 ± 0.01} & 2.15 ± 0.07          & 0.33 ± 0.02          & 6.5 ± 0.02         & 7 ± 6.8            & 0.014 ± 0.009           \\
2.5   & 0.33 ± 0.01          & 2.14 ± 0.26          & 0.36 ± 0.03          & 6 ± 1.01           & 2.2 ± 2.3          & 0.014 ± 0.014           \\
5     & \textbf{0.32 ± 0.01} & \textbf{1.89 ± 0.14} & \textbf{0.31 ± 0.03} & \textbf{5.9 ± 0.3} & \textbf{0.2 ± 0.4} & \textbf{0.002  ± 0.004} \\
\bottomrule
\end{tabular}
\end{table}

Table~\ref{tab:gamma} presents the performance of the proposed UA-PCBF method as the uncertainty weight parameter $\gamma$ varies, considering the values $\gamma \in \{0, 0.5, 1,2.5, 5\}$. This parameter modulates the influence of the predicted uncertainty in human motion on the robot's control decisions.

Taking into account spatial accuracy, the mean hand-TCP distance remains stable across all values of $\gamma$, with the best results (32\,cm) consistently achieved for $\gamma \in \{0, 0.5, 1, 5\}$. This indicates that incorporating uncertainty does not compromise the robot’s ability to maintain proximity to the human hand.

The total TCP path length and average velocity show a clear improvement as $\gamma$ increases. The shortest trajectory (1.89\,m) and lowest average velocity (0.31\,m/s) are achieved at $\gamma = 5$, suggesting that the robot moves more efficiently when uncertainty is fully accounted for.

Similarly, overall task completion time decreases with increasing $\gamma$, reaching a minimum of 5.9\,s at $\gamma = 5$, which is close to the ideal task duration of 5.3\,s. This suggests that higher uncertainty awareness enables the robot to plan more confidently and avoid unnecessary slowdowns.
    
Considering the system's ability to perform the task with maximum safety, the number and magnitude of safety violations drop significantly as $\gamma$ increases. At $\gamma = 5$, the robot incurs only 0.2 violations on average, with a minimal breach magnitude of 2\,cm. This represents an order-of-magnitude improvement over lower $\gamma$ values and highlights the effectiveness of uncertainty-aware control in maintaining safety.

In summary, increasing $\gamma$ enhances both \textit{efficiency} and \textit{safety}, with $\gamma = 5$ yielding the best overall performance. This supports the hypothesis that leveraging uncertainty in human motion prediction leads to more intelligent and fluid robot behavior.

\begin{table}[t]
\centering
\caption{Average values of the considered metrics obtained from the baseline (i.e. CBF and PCBF methods) and UA-PCBF method, also considering the variants with $\gamma = 0$ and $\lambda_p =\lambda_r$. Metrics are presented as mean\,$\pm$\,standard deviation ($\mu\pm\sigma$), where values in bold represent the best results.}
\scriptsize
  \resizebox{\linewidth}{!}{%
\label{tab:baseline}
\begin{tabular}{lcccccc} \toprule
      & \textbf{Hand-TCP }            & \textbf{Trajectory }          & \textbf{Trajectory}           & \textbf{Completion  }       & \textbf{Violation}          & \textbf{Violation }              \\
\textbf{Method} & \textbf{Distance}             & \textbf{Length  }             & \textbf{Velocity }            & \textbf{Time}               & \textbf{Count}              & \textbf{Magnitude}               \\
\midrule
CBF         & 0.36 ± 0.02 & 2.25 ± 0.78 & 0.32 ± 0.1  & 5.9 ± 0.8  & 43.0 ± 14.5   & 0.028 ± 0.004 \\
PCBF        & 0.34 ± 0.01 & 2.26± 0.07  & 0.23 ± 0.01 & 6.7 ± 0.3  & 13.3 ± 13.2 & 0.019 ± 0.013 \\
\hline
\textbf{Ours} ($\lambda_p =\lambda_r$) & \textbf{0.32 ± 0.01} & 2.04 ± 0.07 & 0.33 ± 0.01 & 6.2 ± 0.1  & 0.8 ± 1.8   & 0.002 ± 0.006 \\
\textbf{Ours} ($\gamma = 0$)   & 0.33 ± 0.01 & 2.19 ± 0.15 & 0.34 ± 0.02 & 6.4 ± 0.4  & 7.0 ± 2.5     & 0.020 ± 0.005  \\
\textbf{Ours}     & \textbf{0.32 ± 0.01} & \textbf{1.89 ± 0.14} &\textbf{ 0.31 ± 0.03 }& \textbf{5.9 ± 0.3}  & \textbf{0.2 ± 0.4}   & \textbf{0.002 ± 0.004} \\
\bottomrule
\end{tabular}%
  }
\end{table}
Table~\ref{tab:baseline} compares the full UA-PCBF method (with $\gamma = 5$) against two baselines (CBF and PCBF) and two ablated variants: one with $\gamma = 0$ (no uncertainty awareness in the barrier function) and one with $\lambda_p = \lambda_r$ (no uncertainty awareness in the QP optimization).

UA-PCBF and the $\lambda_p = \lambda_r$ variant achieve the best hand-TCP distance (32\,cm), outperforming both PCBF (34\,cm) and CBF (36\,cm). This suggests that both prediction and uncertainty contribute to maintaining close coordination with the human.
    
UA-PCBF achieves the shortest trajectory (1.89 m) and a velocity (0.31 m/s) close to the ideal (0.3 m/s), indicating efficient and natural motion. In contrast, PCBF and CBF result in longer paths and, in the case of PCBF, a significantly lower velocity (0.23\,m/s), likely due to the purely predictive nature of the method, in contrast with the more reactive CBF/UA-PCBF approaches. In terms of overall execution time, UA-PCBF and CBF both complete the task in 5.9\,s, but UA-PCBF does so with far fewer safety violations, highlighting its superior balance between speed and safety.
    
Considering the comparison in terms of safety compliance, UA-PCBF achieves the lowest violation count (0.2) and smallest violation magnitude (0.2\,cm), significantly outperforming PCBF (13.3 violations) and CBF (43 violations). The $\gamma = 0$ variant performs poorly in this regard, confirming the critical role of uncertainty modeling. Interestingly, the $\lambda_p = \lambda_r$ variant also performs well in safety, suggesting that even without prediction-based relaxation or tightening of optimization constraints, uncertainty-aware control can significantly enhance safety.

These results demonstrate that UA-PCBF enables robots to navigate the trade-off between responsiveness and safety more effectively than existing methods. By incorporating uncertainty in human motion prediction, robots can plan more confidently, avoid unnecessary braking, and maintain safe, efficient, and fluid interactions in shared workspaces.

\subsection{Human operator experiment}
\label{sec:operator_results}
\begin{table}[ht]
\centering
\caption{Average values of the considered metrics obtained from the baseline (i.e. CBF and PCBF methods) and UA-PCBF method during interaction with human operator. Metrics are presented as mean\,$\pm$\,standard deviation ($\mu\pm\sigma$), where values in bold represent the best results.}
\scriptsize
  \resizebox{\linewidth}{!}{%
\label{tab:operator}
\begin{tabular}{lcccccc} \toprule
               & \textbf{Hand-TCP } & \textbf{Trajectory } & \textbf{Trajectory} & \textbf{Completion} & \textbf{Violation} & \textbf{Violation } \\
\textbf{Method} & \textbf{Distance} & \textbf{Length  }  & \textbf{Velocity }    & \textbf{Time}         & \textbf{Count}  & \textbf{Magnitude}   \\
\midrule
CBF          & \textbf{0.30 ± 0.01} & 2.75 ± 0.05 & 0.97 ± 0.01  & 5.84 ± 0.4  & 94.3 ± 5.7   & 0.023 ± 0.001 \\
PCBF          & 0.32 ± 0.01  & 2.5± 0.14  & 0.44 ± 0.11 & 5.65 ± 0.8  & 34.3 ± 24 & 0.036 ± 0.006 \\
\textbf{UA-PCBF}   & 0.34 ± 0.03 & \textbf{2.18 ± 0.17} & \textbf{0.41 ± 0.04} & \textbf{5.34 ± 1.1}  & \textbf{2 ± 3.5 }  & \textbf{0.012 ± 0.01} \\
\bottomrule
\end{tabular}%
  }
\end{table}

\Cref{tab:operator} presents the values of the metrics obtained in the experiment with a human operator, also considering the two baseline methods CBF and PCBF as in the previous mockup experiments. 

CBF achieves the lowest average distance (30\,cm), followed by PCBF (32\,cm) and UA-PCBF (34\,cm). This suggests that CBF maintains closer proximity to the human hand, but this may come at the cost of safety. In fact, CBF’s high velocity (0.97\,m/s) reflects aggressive behavior, potentially unsafe considering the close proximity to humans. Instead UA-PCBF achieves the lowest average velocity (0.41\,m/s), lower than those obtained from PCBF (0.44\,m/s). While this might suggest an overly conservative behavior of UA-PCBFs, it is not reflected in a reduction in task execution efficiency.

In fact, UA-PCBF yields the shortest trajectory (2.18\,m), slightly better than PCBF (2.5\,m), and significantly shorter than CBF (2.75\,m). This indicates more efficient motion planning under UA-PCBF, an aspect that is also confirmed by the task completion times. UA-PCBF completes the task in 5.34\,s on average, faster than PCBF (5.65\,s) and comparable to CBF (5.84\,s), despite its more conservative velocity.

If we consider the performance in terms of safety compliance, A-PCBF reduces safety violations to just 2 on average, compared to 34.3 for PCBF and 94.3 for CBF. Also, UA-PCBF also achieves the lowest breach depth (1.2\,cm), outperforming PCBF (3.6\,cm) and CBF (2.3\,cm).

The results confirm that UA-PCBF maintains its advantages even in dynamic, human-in-the-loop scenarios. It balances efficiency and safety better than both CBF and PCBF, adapting to unpredictable human motion while minimizing risky behavior. The UA-PCBF method demonstrated slightly reduced performance in the human operator scenario compared to the hand mockup experiment, with a higher average violation count (2 vs. 0.2) and greater violation magnitude (1.2\,cm vs. 0.2\,cm), reflecting the increased unpredictability of real human motion. 
The degradation in performance compared to the mockup experiment is expected and underscores the importance of robust uncertainty modeling.

\section{Limitations}\label{sec:limitations}



Our study exhibits several limitations. First, the experimental scenarios remain highly simplified: initial validation relied on a static hand model executing repeatable trajectories, and even the human‐in‐the‐loop handover task involved only a single operator following prescribed motions, rather than a broad set of unstructured collaborative activities. Second, no formal user testing or usability studies were conducted, so we lack data on how real workers perceive system responsiveness and cognitive load during prolonged interaction. Finally, our safety model employs geometric approximations of the bounding region for the hand and for the robot’s end‐effector and links. 

\section{Conclusions}\label{sec:conclusions}



This work introduced Uncertainty-Aware Predictive Control Barrier Functions (UA-PCBFs), a novel framework for enabling safe and efficient human–robot interaction in shared workspaces. 
We integrate probabilistic human motion prediction with the formal safety guarantees of Control Barrier Functions, allowing robots to reason not only about the most likely future human states but also about the uncertainty surrounding those predictions. 
This allows us to have a dynamic adjustment of the robot's planning and control, based on the uncertainty estimation of the human motion.

Through a series of controlled experiments involving a robotic manipulator interacting with a human hand model, we demonstrated that UA-PCBFs outperform both classical CBFs and prediction-augmented PCBFs across a range of metrics. Specifically, UA-PCBFs achieved significant reductions in safety violations—both in frequency and magnitude—while maintaining or improving task efficiency and spatial coordination. These results validate the core hypothesis that uncertainty-aware planning leads to more fluid and intelligent robot behavior in collaborative settings.

Future works will extend the current framework to more complex tasks, such as those of Industry 5.0 and Human-Robot Collaboration, with an extensive user study to validate both usability and impact on people. 




\section*{Credit authorship contribution statement}


\textbf{Lorenzo Busellato:} Methodology, Formal analysis, Software, Writing-Original draft preparation.
\textbf{Federico Cunico:} Methodology, Data curation, Software, Writing-Original draft preparation.
\textbf{Diego Dall'Alba:} Investigation, Visualization, Writing-Reviewing and Editing.
\textbf{Marco Emporio:} Validation, Resources.
\textbf{Andrea Giachetti:} Supervision, Resources.
\textbf{Riccardo Muradore:} Conceptualization, Supervision, Writing-Reviewing.
\textbf{Marco Cristani:} Conceptualization, Funding acquisition, Supervision, Writing-Reviewing and Editing.

\section*{Declaration of competing interest}

The authors declare that they have no known competing financial interests or personal relationships that could have appeared to influence the work reported in this paper.
\section*{Acknowledgments}
This study was carried out within the MICS (Made in Italy – Circular and Sustainable) Extended Partnership and received funding from Next-Generation EU (Italian PNRR – M4 C2, Invest 1.3 – D.D. 1551.11-10-2022, PE00000004). CUP MICS D43C22003120001 - Cascade funding project CollaborICE.

This manuscript reflects only the Authors’ views and opinions, neither the European Union nor the European Commission can be considered responsible for them.

\bibliographystyle{elsarticle-num}
\bibliography{biblio}
\end{document}